\documentclass[runningheads]{llncs}
\usepackage[T1]{fontenc}
\usepackage{graphicx,verbatim}
\usepackage{enumitem}
\usepackage{booktabs}
\usepackage{multirow}
\usepackage{siunitx}
\usepackage{amsmath,amssymb,amsfonts}

\usepackage[acronym,shortcuts]{glossaries}
\newacronym{MSE}{MSE}{mean squared error}
\newacronym{SSIM}{SSIM}{structural similarity index measure}
\newacronym{V1}{V1}{primary visual cortex}
\glsdisablehyper  

\usepackage[table]{xcolor}         

\begin{document}
%
\title{Percept-Aware Surgical Planning for Visual Cortical Prostheses with Vascular Avoidance}
\titlerunning{Percept-Aware Surgical Planning for Visual Cortical Prostheses}
%

\author{Galen Pogoncheff\inst{1} \and
Alvin Wang\inst{2} \and
Jacob Granley\inst{1} \and
Michael Beyeler\inst{1,3}
}
\authorrunning{G. Pogoncheff et al.}
%
\institute{Department of Computer Science, University of California Santa, Barbara \and
School of Computer Science, Carnegie Mellon University \and
Department of Psychological \& Brain Sciences, University of California, Santa Barbara
}

\maketitle              
\begin{abstract}
Cortical visual prostheses aim to restore sight by electrically stimulating neurons in early visual cortex (V1). With the emergence of high-density and flexible neural interfaces, electrode placement within three-dimensional cortex has become a critical surgical planning problem. Existing strategies emphasize visual field coverage and anatomical heuristics but do not directly optimize predicted perceptual outcomes under safety constraints.
We present a percept-aware framework for surgical planning of cortical visual prostheses that formulates electrode placement as a constrained optimization problem in anatomical space. Electrode coordinates are treated as learnable parameters and optimized end-to-end using a differentiable forward model of prosthetic vision. The objective minimizes task-level perceptual error while incorporating vascular avoidance and gray matter feasibility constraints.
Evaluated on simulated reading and natural image tasks using realistic folded cortical geometry (FreeSurfer \emph{fsaverage}), percept-aware optimization consistently improves reconstruction fidelity relative to coverage-based placement strategies. Importantly, vascular safety constraints eliminate margin violations while preserving perceptual performance. The framework further enables co-optimization of multi-electrode thread configurations under fixed insertion budgets.
These results demonstrate how differentiable percept models can inform anatomically grounded, safety-aware computer-assisted planning for cortical neural interfaces and provide a foundation for optimizing next-generation visual prostheses.

\end{abstract}
\section{Introduction}

Cortical visual neuroprostheses aim to restore sight by electrically stimulating neurons in early visual cortex, eliciting visual percepts known as phosphenes~\cite{fernandez_development_2018}. With the emergence of high-density and flexible neural interfaces~\cite{musk_integrated_2019,merken_thin_2022}, surgeons now face an increasingly critical question: where should electrodes be placed within three-dimensional cortex to maximize functional visual benefit?

Large-scale stimulation of \ac{V1} has demonstrated phos\-phene-based pattern vision in non-human primates~\cite{chen_shape_2020}, but translating such systems to humans is complicated by greater gyrification and substantial inter-individual variability~\cite{benson_variability_2022}.
Importantly, the functional organization of early visual cortex is largely predictable from cortical anatomy~\cite{benson2018bayesian}, which is especially critical in blind patients where visually-driven fMRI mapping is not feasible~\cite{striem-amit_functional_2015}. 
Electrode design and placement, together with cortical geometry, ultimately determine the structure of the resulting phosphene map.

Current placement strategies rely on anatomical landmarks and retinotopic maps~\cite{fernandez2021visual}, targeting foveal representation while avoiding vasculature. 
Although safety-aware, these approaches remain geometric and coverage-driven rather than optimized for predicted perceptual fidelity and functional utility. 
Prior work has optimized stimulation for fixed implants or arranged electrodes to maximize visual field coverage~\cite{bruce_greedy_2022,hoof_optimal_2024}, but perceptual objectives and surgical constraints have not been jointly optimized within a unified planning framework.

Computational modeling has shown that both stimulation parameters and electrode location shape phosphene structure~\cite{beyeler_model_2019,grinten_biologically_2022,fine_virtual_2024}. Prior work has either optimized stimulation for fixed implants or arranged electrodes to maximize visual field coverage~\cite{bruce_greedy_2022,hoof_optimal_2024}. However, perceptual objectives and surgical constraints have not been jointly optimized within a unified planning framework.

We introduce a model-driven approach for pre-operative planning that directly optimizes cortical electrode placement for perceptual fidelity under anatomical constraints. 
Electrode coordinates are treated as learnable parameters and optimized end-to-end using differentiable percept models. 
The framework incorporates retinotopic mappings and vascular avoidance, enabling percept-aware surgical optimization on realistic folded cortex.
We demonstrate the method on the widely used FreeSurfer \emph{fsaverage} cortical template, allowing optimization over folded cortical geometry in a consistent anatomical reference space.

\textbf{The main contributions of our work include} \textbf{(1)} a percept-aware optimization framework for three-dimensional cortical electrode placement that directly minimizes predicted task-relevant perceptual error, \textbf{(2)} explicit integration of vascular avoidance and anatomical feasibility constraints into the placement optimization, and \textbf{(3)} demonstration that optimized configurations can reduce cortical insertions while preserving functional performance.

\section{Related Work}

Research on visual neuroprostheses has largely progressed along two directions: (i) optimizing stimulation strategies for fixed implant geometries, and (ii) optimizing implant placement based on visual field coverage. Our work connects these lines by directly optimizing cortical electrode placement using a perceptual objective under anatomical constraints.

Prior work has focused on selecting stimulation parameters for a given electrode array, including optimization of current amplitudes and electrode combinations \cite{shah_optimization_2019,bratu_graph-based_2020}, hybrid calibration of electrode sensitivity profiles \cite{granley_human---loop_2023}, and encoding frameworks that map images to stimulation patterns through learned transformations \cite{granley_hybrid_2022,de_ruyter_van_steveninck_end--end_2022}. These approaches demonstrate that model-based optimization improves perceptual outcomes, but they assume implant geometry and location are fixed and therefore do not address surgical placement.

A complementary line of work examines how implant location affects perceptual structure. In retinal implants, electrode position strongly influences phos\-phene geometry due to axonal organization \cite{beyeler_model_2019}, motivating model-based selection of intraocular implant configurations \cite{beyeler_model-based_2019}. Subsequent strategies optimized electrode arrangements to improve visual field tiling rather than retinal surface coverage \cite{bruce_greedy_2022}. In cortex, retinotopic maps have been used to position electrodes to maximize projected visual field coverage \cite{hoof_optimal_2024}. However, these methods optimize geometric coverage rather than task-level perceptual fidelity and typically rely on simplified percept approximations.

\emph{Literature Gap:}
Existing work either optimizes stimulation for fixed implants or optimizes placement using coverage-based objectives. Perceptual fidelity, task-relevant utility, and surgical constraints have not been jointly optimized within a planning framework. In contrast, we formulate cortical electrode placement as a percept-aware optimization problem in 3D brain space, integrating retinotopy and vascular constraints to support pre-operative surgical planning.

\begin{figure}[t!]
    \centering
    \includegraphics[width=0.9\textwidth]{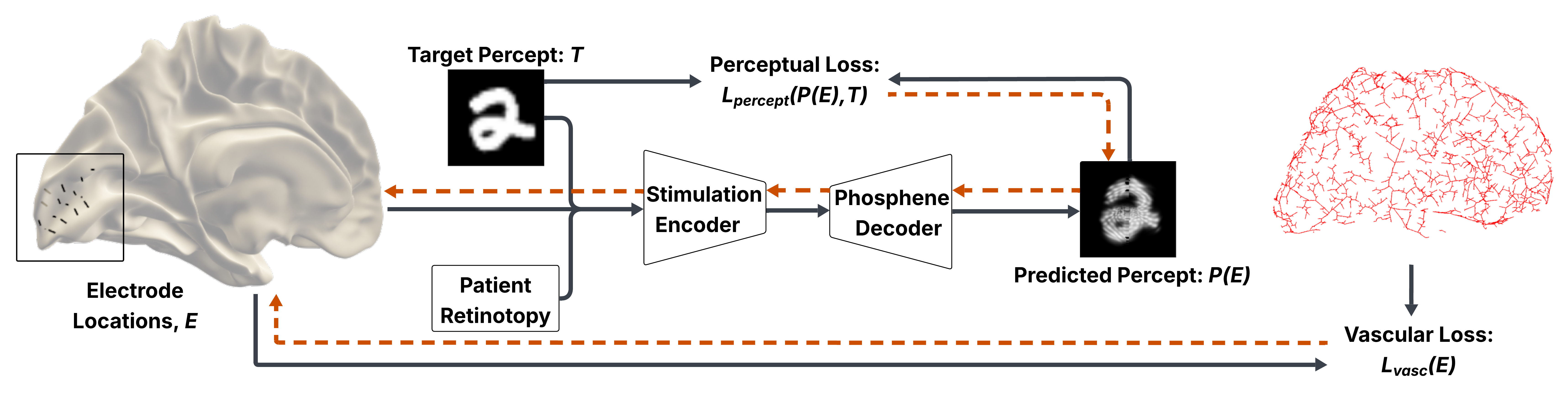}
    \caption{Percept-aware optimization framework. Given 3D electrode coordinates on FreeSurfer \emph{fsaverage} anatomy, target percepts, and patient retinotopy, a differentiable model of cortical prosthetic vision predicts elicited percepts. Electrode locations are iteratively updated to minimize perceptual error while enforcing vascular safety constraints. Solid arrows: forward simulation pathway. Dashed arrows: gradient signals used to update electrode positions.} \label{fig:system_diagram}
\end{figure}

\section{Methods}

\subsection{Surgical Planning Formulation}

We formulate cortical electrode placement as a constrained surgical optimization problem in three-dimensional brain space. Let $\Omega \subset \mathbb{R}^3$ denote the feasible cortical region corresponding to gray matter in early visual cortex. Electrode locations are parameterized as $E = \{\mathbf{p}_e\}_{e=1}^N$, where $\mathbf{p}_e = (x_e,y_e,z_e) \in \Omega$ represents the cortical coordinate of electrode $e$, and $N$ is the fixed number of stimulation sites determined by device design.

Given (i) a retinotopic mapping from cortical coordinates to visual field locations, (ii) a vascular map defining prohibited regions, and (iii) a distribution of task-relevant target percepts $T \sim \mathcal{D}$, the objective is to determine electrode locations that maximize predicted perceptual fidelity while satisfying anatomical safety constraints (Fig.~\ref{fig:system_diagram}). 

Electrode placement is optimized within a differentiable forward model of cortical prosthetic vision, allowing predicted percepts to be expressed as differentiable functions of electrode coordinates. This enables gradient-based optimization directly in anatomical space.

We solve for optimal placement via the following multi-objective formulation:
$E^* = \arg\min_{E \subset \Omega}
\;
\mathbb{E}_{T \sim \mathcal{D}}
\left[
\mathcal{L}_{\text{percept}}(P(E), T)
\right]
+
\lambda_{\text{vasc}} \mathcal{L}_{\text{vasc}}(E)
+
\lambda_{\text{cortex}} \mathcal{L}_{\text{cortex}}(E),$
where $\mathcal{L}_{\text{percept}}$ measures task-level perceptual error, $\mathcal{L}_{\text{vasc}}$ penalizes violations of vascular safety margins, and $\mathcal{L}_{\text{cortex}}$ enforces anatomical feasibility. Scalar weights $\lambda_{\text{vasc}}$ and $\lambda_{\text{cortex}}$ control the trade-off between functional performance and surgical safety.

\subsection{Anatomical Representation}

Electrode placement was performed on the FreeSurfer \emph{fsaverage} cortical template. The template provides a folded gray matter surface mesh and volumetric representation, enabling optimization directly within realistic cortical geometry.

The feasible region $\Omega$ was defined as cortical gray matter within early visual cortex. Gray matter localization was obtained from the template segmentation and used to constrain electrode coordinates to anatomically plausible locations. Although laminar boundaries were not explicitly modeled, confinement to gray matter serves as a proxy for targeting layer 4, the typical stimulation depth for intracortical visual prostheses.

Retinotopic mappings were obtained from the Benson et al.~\cite{benson2018bayesian} Bayesian retinotopy prior defined in \emph{fsaverage} surface space. The cerebrovascular atlas~\cite{ii2020multiscale} was registered to the same coordinate system, enabling direct computation of electrode-to-vessel distances. All anatomical constraints and percept simulations were therefore evaluated in a common neuroanatomical reference frame.

\subsection{Perceptual Forward Model}

We define a differentiable forward model that maps cortical electrode coordinates to predicted visual percepts. The model consists of two stages: (i) a retinotopic mapping from cortical space to visual field coordinates, and (ii) a phosphene generation model that predicts the elicited percept.

\emph{Retinotopic Mapping:}
For an electrode at cortical location $\mathbf{p}_e$, its visual field coordinate $\mathbf{s}_e$ is estimated by distance-weighted interpolation over the $k$ nearest retinotopic sites, $\mathcal{N}_k(\mathbf{p}_e)$, each of which is mapped to a visual field location $\mathbf{v}_j$:
\begin{equation}
\mathbf{s}_e =
\frac{\sum_{j \in \mathcal{N}_k(\mathbf{p}_e)}
w_{ej}\mathbf{v}_j}
{\sum_{j \in \mathcal{N}_k(\mathbf{p}_e)} w_{ej}},
\quad
w_{ej} = \frac{1}{\|\mathbf{p}_e - \mathbf{c}_j\|_2}.
\end{equation}

\emph{Phosphene Generation:}
Stimulation amplitude $a_e$ is sampled from the target at $\mathbf{s}_e$, and the percept is modeled as a sum of isotropic Gaussians with spread $\rho$:
$P(x,y)=\sum_{e=1}^{N} a_e \exp\!\left(-\frac{(x-\mathbf{s}_{e, x})^2+(y-\mathbf{s}_{e, y})^2}{2\rho^2}\right)$.

This formulation provides a differentiable mapping from cortical electrode coordinates to image-space percepts, enabling gradient-based optimization in anatomical space. While Gaussian phosphenes with linear superposition provide a tractable approximation consistent with prior modeling work~\cite{fernandez2021visual,grinten_biologically_2022}, the framework readily generalizes to alternative, differentiable percept models.

\emph{Perceptual Objective:}
Perceptual fidelity is measured using a foveally weighted mean squared error: $\mathcal{L}_{\text{percept}}(P,T) = \sum_{(x,y)} w(x,y)\left(P(x,y)-T(x,y)\right)^2$,
where $w(x,y)$ increases toward the fovea to reflect the higher functional importance of central vision in many tasks.

\subsection{Anatomical and Safety Constraints}

Electrode placement is subject to anatomical feasibility and vascular safety constraints. These are incorporated as differentiable penalty terms within the optimization objective.

Avoidance of cortical vasculature is a primary safety consideration in intracortical implantation. Prior studies report vascular disruption extending up to \SI{300}{\micro\meter} from insertion sites~\cite{bjornsson2006effects}, motivating an explicit safety margin.
Let $\mathcal{V}$ denote the set of vascular structures and $d(\mathbf{p}_e, \mathcal{V})$ the minimum Euclidean distance between electrode location $\mathbf{p}_e$ and the vasculature. We define a hinge-style penalty $\mathcal{L}_{\text{vasc}}(E) = \frac{1}{|E|} \sum_{e \in E}$ if $d(\mathbf{p}_e, \mathcal{V}) \le \tau$, 
where $\tau = \SI{300}{\micro\meter}$ defines the safety margin,
and 0 otherwise.
This produces strong gradients near vascular boundaries while allowing unconstrained optimization in safe regions.

Intracortical stimulation targets neurons within cortical gray matter. To ensure anatomically plausible placements, we penalize electrode coordinates outside the gray matter mask. Let $d_{\text{gm}}(\mathbf{p}_e)$ denote the signed distance to gray matter. We define $\mathcal{L}_{\text{cortex}}(E) = \frac{1}{|E|} \sum_{e \in E} d_{\text{gm}}(\mathbf{p}_e)^2$.

Together, these terms embed surgical feasibility directly into the differentiable planning framework, enabling trade-offs between functional performance and safety to be explored continuously during optimization.

\subsection{Optimization Procedure}

Optimization was performed using gradient-based updates in anatomical space. Because the forward percept model and constraint terms are differentiable w.r.t. electrode coordinates, gradients were computed via automatic differentiation. During optimization, vascular and gray matter penalties were jointly applied with perceptual loss, allowing electrode locations to be continuously adjusted to balance functional performance and surgical safety.

Electrode coordinates were initialized uniformly within the feasible cortical region $\Omega$. We used the Adam optimizer with fixed learning rate and optimized electrode positions until convergence of the objective on training data. Experiments were performed in Python using TensorFlow. Optimization for a single configuration required less than 10 min. on a NVIDIA RTX 3090 GPU.

\subsection{Experimental Protocol}

\emph{Datasets:}
We evaluated placement strategies using two image datasets representing distinct functional goals of visual prostheses: handwritten digits from MNIST~\cite{deng2012mnist}, approximating structured symbol recognition (e.g., reading), and natural images from CIFAR-10~\cite{krizhevsky2009learning}, representing more complex visual scenes.

\emph{Experimental Configurations:}
Experiments were conducted across electrode counts, $N \in \{64-1024\}$, and phosphene spreads, $\rho \in \{500-1500\}$ \SI{}{\micro\meter}. For each configuration, optimization was performed over $3$ random initializations.

\emph{Baselines:}
We compared the proposed percept-aware optimization against two placement strategies:
(i) \emph{Visual Field Tiling}, which uniformly tiles the visual field, and
(ii) \emph{Visual Field Coverage}, which optimizes electrode positions to maximize coverage of the visual field represented in the training images~\cite{hoof_optimal_2024}.

\emph{Evaluation Metrics:}
Perceptual fidelity was quantified using \ac{SSIM} and \ac{MSE} between target images and simulated percepts, along with downstream classification accuracy. SSIM captures structural and luminance consistency, while MSE provides a global reconstruction error measure. Learned perceptual metrics were not used because prosthetic percepts deviate substantially from natural image statistics, particularly at low electrode counts~\cite{granley_hybrid_2022}. To assess task-specific functional utility (symbol and object identification), we trained image classifiers (ResNet-50 \cite{he2016deep}) on target images and evaluated performance on simulated phosphenes \cite{granley_human---loop_2023}.

\emph{Statistical Analysis:}
Results were aggregated across experimental configurations and random initializations. Statistical comparisons between methods were performed using Wilcoxon signed-rank tests with significance threshold $p \le 0.01$.

\begin{table}[b!]
    \caption{Relative change in MSE, SSIM, and downstream classification accuracy (Acc.) of the proposed method versus baselines. Values are median percent difference [IQR] across electrode counts and phosphene spreads ($\rho$). * indicates $p \le 0.01$.}
    \centering
    \setlength{\tabcolsep}{2pt}
    \begin{tabular}{l c c c c}
    \toprule
        Baseline & Dataset & $\Delta$ MSE $(\%)$ $\downarrow$ & $\Delta$ SSIM $(\%)$ $\uparrow$ & $\Delta$ Acc. $(\%)$ $\uparrow$ \\
        \midrule
        VF Tiling & \multirow{2}{*}{MNIST} & -67.7* [-73.6, -47.2] & 11.1* [7.8, 16.7] & 62.6* [28.9, 119.9] \\
        VF Coverage &  & -33.4* [-42.0, -16.3] & 4.7* [3.1, 7.3] & 22.4* [7.8, 34.7] \\
        \midrule
        VF Tiling & \multirow{2}{*}{CIFAR-10} & -58.4* [-87.4, -24.1] & 57.4* [35.3, 102.2] & 10.2* [5.5, 29.2] \\
        VF Coverage &  & 16.0 [-25.2, 39.1] & -10.6 [-24.7, 0.2] & -2.2 [-9.1, 6.25] \\
    \bottomrule
    \end{tabular}
    \label{tab:baseline_comarisons}
\end{table}

\section{Results}

\subsection{Percept-Aware Optimization Improves Functional Fidelity}

We first evaluated placement optimization without anatomical constraints to isolate functional performance. Across electrode counts and phosphene spread parameters, percept-aware optimization consistently reduced reconstruction error relative to both Visual Field Tiling and Visual Field Coverage.

\begin{figure}[t!]
    \centering
    \includegraphics[width=\textwidth]{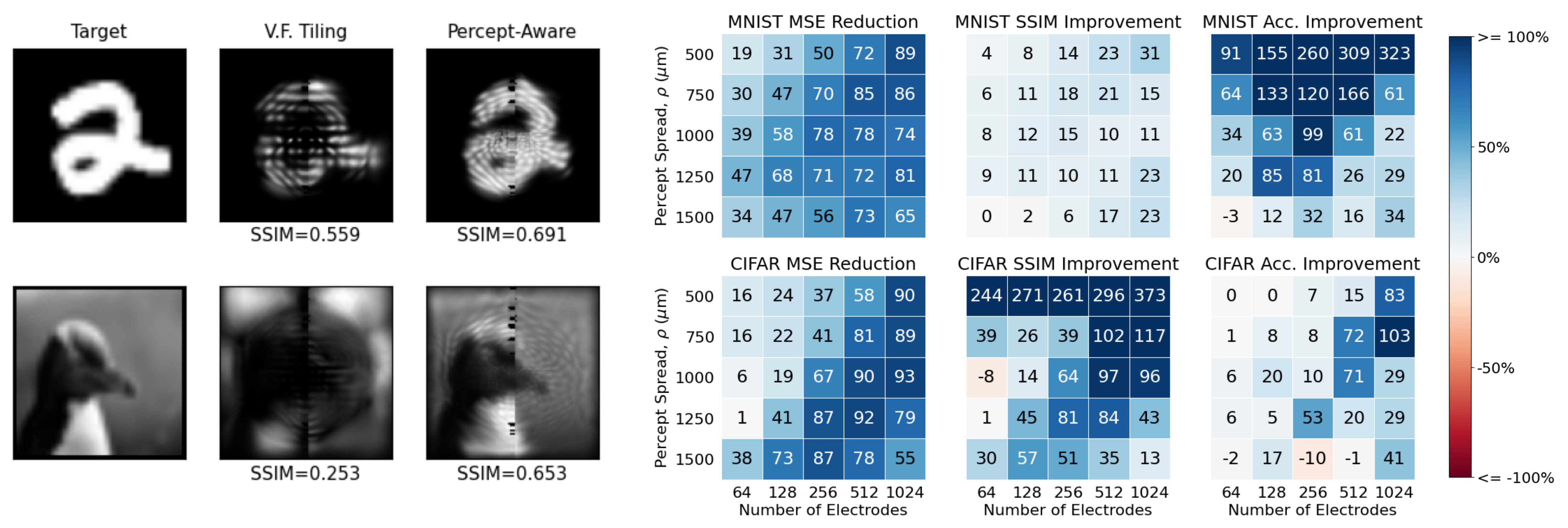}
    \caption{Comparing percept-aware electrode optimization and Visual Field Tiling. \textit{Left}: Simulated phosphenes for reading (MNIST) and natural image (CIFAR-10) tasks. \textit{Right}: Relative improvements in perceptual fidelity across experimental configurations.
    } \label{fig:comparison_to_baseline}
\end{figure}

On MNIST, percept-aware optimization reduced median MSE by up to 67.7\% relative to tiling and 33.4\% relative to coverage, increased downstream classification accuracy by 62.6\% and 22.4\%, respectively, and yielded corresponding improvements in SSIM (Table~\ref{tab:baseline_comarisons}). On CIFAR-10, percept-aware placement substantially outperformed tiling and achieved performance comparable to coverage-based optimization. The latter is expected because CIFAR-10 images typically occupy the full visual field, aligning coverage objectives with reconstruction fidelity.
Qualitative examples (Fig.~\ref{fig:comparison_to_baseline}) show clearer symbol structure and improved spatial coherence under percept-aware placement.

\begin{figure}[b!]
\centering
\includegraphics[width=0.9\textwidth]{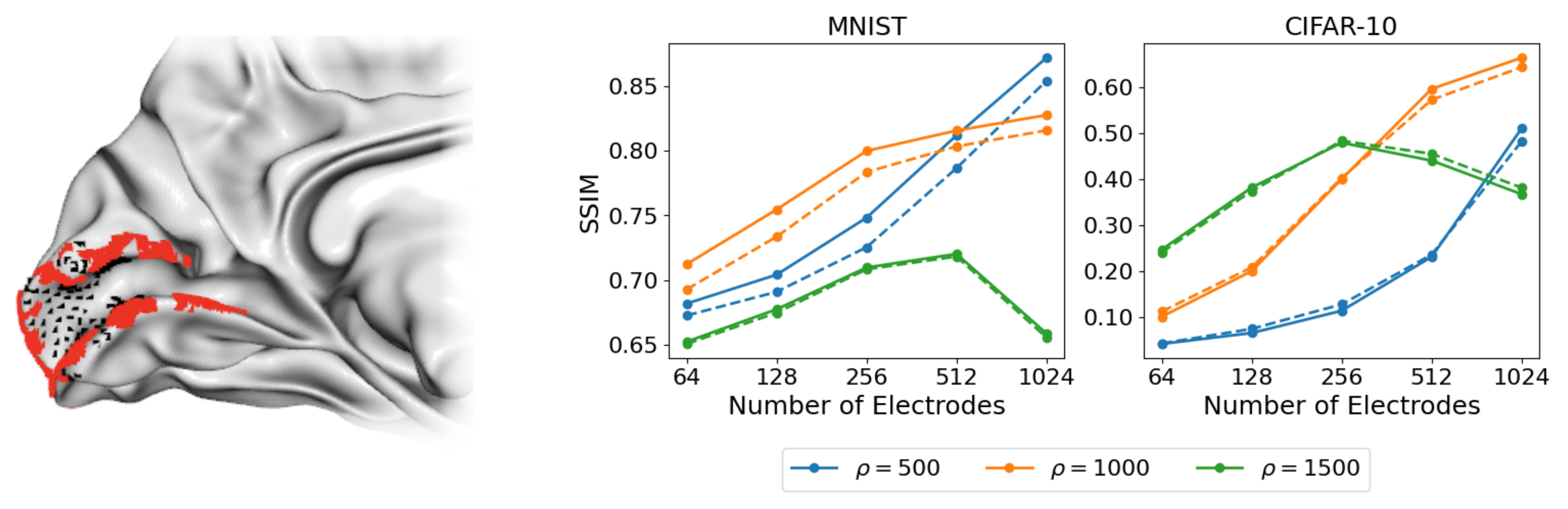}
\caption{Safety-aware electrode optimization with vascular constraints. \textit{Left}: Cortical surface (\emph{fsaverage} left occipital) with high-resolution vascular map (red, displayed in V1 only) and optimized electrode locations (black). \textit{Right}: Percept SSIM scores without (solid lines) and with (dashed lines) vascular avoidance.} \label{fig:vasc}
\end{figure}

\subsection{Safety-Aware Optimization Preserves Perceptual Performance}

We next incorporated vascular avoidance into the optimization objective. Without safety constraints, a large fraction of electrodes violated the \SI{300}{\micro\meter} safety margin. Incorporating the vascular penalty eliminated all margin violations.

Perceptual quality remained largely unchanged. Across configurations, SSIM decreased by only 1.7\% on MNIST and 4.4\% on CIFAR-10 relative to unconstrained optimization (Fig.~\ref{fig:vasc}), demonstrating maintenance of functional performance while satisfying clinically-motivated vascular safety constraints.

\subsection{Device Architecture Co-Optimization with Threaded Arrays}

We extended the framework to optimize multi-electrode threads~\cite{musk_integrated_2019} under a fixed number of cortical insertions. In this setting, entry location and insertion trajectory were jointly optimized within the same percept-aware objective.

\begin{figure}[tb!]
    \centering
    \includegraphics[width=.9\textwidth]{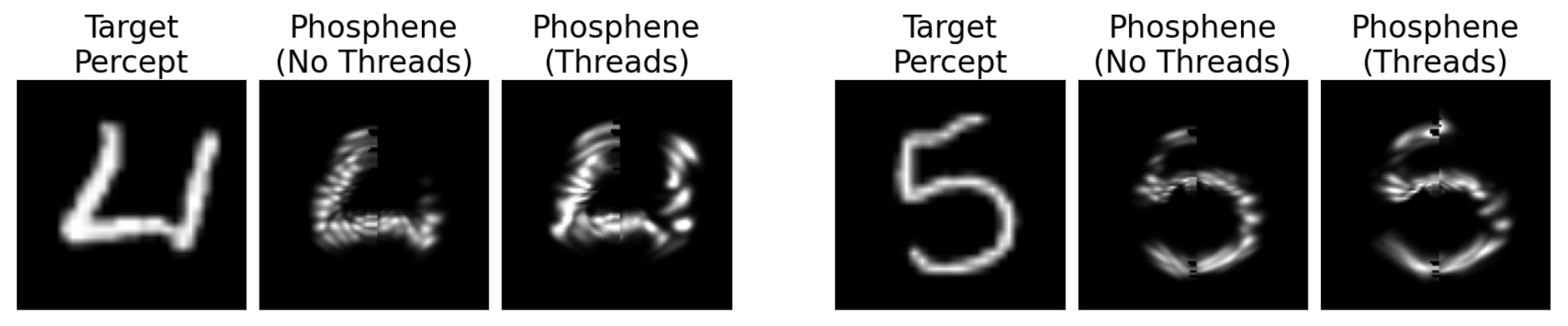}
    \caption{Simulated phosphenes using percept-aware optimization with and without multi-electrode threads under a fixed number of cortical insertions ($128$). Threaded designs improve percept quality without increasing number of cortical insertions.} \label{fig:mnist_threads}
\end{figure}

Under a fixed insertion budget ($N_{\text{insert}} = 128$), threaded configurations improved perceptual fidelity relative to single-electrode placements (Fig.~\ref{fig:mnist_threads}). These results highlight the framework’s ability to co-optimize surgical constraints and device architecture, enabling quantitative exploration of device design trade-offs. 

\section{Discussion}

We present a percept-aware surgical planning framework for cortical visual prostheses that directly optimizes electrode placement in three-dimensional brain space under anatomical constraints. By integrating differentiable percept modeling with realistic cortical geometry and vascular avoidance, the proposed approach aligns surgical decision-making with predicted functional outcomes.

Across simulated reading and natural image tasks, percept-aware placement consistently improved reconstruction fidelity relative to coverage-based strategies. Importantly, incorporating vascular safety constraints preserved perceptual performance while eliminating margin violations, demonstrating that functional optimization and surgical safety can be addressed jointly within a unified framework. The extension to multi-electrode threads further illustrates how the method enables co-optimization of device design and surgical constraints.

Several limitations remain. First, experiments were conducted on the widely used \emph{fsaverage} template rather than subject-specific anatomy. However, the framework is fully compatible with patient-specific cortical reconstructions, retinotopic mapping (e.g., fMRI), and angiographic data. Second, the perceptual forward model uses simplified Gaussian phosphenes and linear superposition. Future work will incorporate more detailed biophysical models.

Overall, this work introduces perceptual objectives into computer-assisted planning for cortical neural interfaces and establishes a framework for anatomically grounded, safety-aware optimization of next-generation visual prostheses.

%
%
%
%

\bibliographystyle{splncs04}
\bibliography{references}

\end{document}